\documentclass[runningheads]{llncs}

\usepackage[final,year=2026]{accv}

\usepackage{lmodern}

\usepackage{accvabbrv}

\usepackage{graphicx}
\usepackage{booktabs}
\usepackage{placeins}
\usepackage{array}
\newcolumntype{L}[1]{>{\raggedright\arraybackslash}p{#1}}

\usepackage[accsupp]{axessibility}  

\usepackage[breaklinks]{hyperref}

\graphicspath{{figures/}}

\providecommand{\papertitle}{Plausible but Unimplantable: A Trustworthiness Audit of
Shape Completion Foundation Models for Cranial Defect Reconstruction}

\makeatletter
\@ifundefined{ifpaperinbody}{\newif\ifpaperinbody}{}
\ifx\@nodocument\relax\paperinbodytrue\else\paperinbodyfalse\fi
\makeatother

\ifpaperinbody

\begin{abstract}
A patient who has lost part of the skull vault needs an implant that fits the defect
exactly, and such implants are still drawn by hand in CAD by specialists, which makes
each case slow and expensive. The task can be considered as a shape completion problem,
so pretrained completion models, and now the first shape completion foundation models,
look like a way to remove that manual step. We audited three publicly released pretrained
shape completion models, two of them foundation models, applied zero-shot to 239 cranial defects,
some carved from public skull datasets and some real post-craniectomy cases. Each output was
compared with simply returning the patient's own remaining bone, and tested for whether the
predicted material could be placed without intersecting the bone that is still there. On the 210 carved defects none of the three improved on doing nothing, and on
the 29 real cases only one did so beyond chance. Yet not one output, from any model in any
configuration, was implantable: none managed both to stay clear of the bone the patient
still has and to place more material than doing nothing, where a task-specific volumetric model, trained for this task and selected on
these cases, was implantable on all but one. The output that most convincingly resembles a
skull is no more placeable than the rest, and at these point
densities implant thickness cannot be read off the emitted points at all, which is a property
of the output representation rather than of any one model. On this evidence the shape completion
foundation models we could obtain cannot be trusted and, used as released, are not usable in
clinical practice. Substantial research
towards trustworthy foundation models for shape completion is still needed.

\end{abstract}

\fi

\begin{document}

\title{\papertitle}
\titlerunning{Plausible but Unimplantable}

\author{Marek Wodzinski\inst{1,2}\orcidID{0000-0002-8076-6246} \and
Kamil Kwarciak\inst{1}\orcidID{0000-0002-1392-4291} \and
Daria Hemmerling\inst{1}\orcidID{0000-0002-2193-7690} \and
Mateusz~Daniol\inst{1}\orcidID{0000-0003-2363-7912}}
\authorrunning{M. Wodzinski et al.}
\institute{AGH University of Krakow, Krakow, Poland \and
Sano Centre for Computational Medicine, Krakow, Poland\\
\email{wodzinski@agh.edu.pl}}

\maketitle

{\renewcommand{\thefootnote}{\fnsymbol{footnote}}%
\footnotetext[1]{Accepted at the Workshop on Trustworthy Foundation Models in Medical Imaging
(TrustFMI), ACCV 2026. This is the authors' accepted manuscript; the Version of Record will appear
in the ACCV 2026 workshop proceedings.}}

\section{Introduction}
\label{sec:intro}

After a decompressive craniectomy, a head injury or a tumour resection that takes bone, a patient
is left with an opening in the skull vault, closed with an implant shaped to that patient's
anatomy~\cite{autoimplant2020,autoimplant2021}. The implant should protect the brain and restore
the original shape of the skull, and whether it fits is settled in the operating
room~\cite{mug500,li2021cranioplasty}.

Such implants are designed by hand, in CAD, by specialists, which makes each case slow and
expensive and limits how many patients can be offered one. Automation has been pursued for a
decade, mostly as volumetric completion of the missing bone, and networks trained for that task now
do well on public benchmarks~\cite{autoimplant2020,autoimplant2021,li2021cranioplasty}, most
recently point-cloud diffusion and flow
matching~\cite{friedrich2023,kwarciak2026pcfm,kwarciak2026ted}. Every one of them has to be trained
on data for the task, and that is the cost a foundation model promises to remove.

Foundation models are becoming increasingly popular in medical
imaging~\cite{moor2023,zhang2024foundation}, and the first of them aimed at shape completion are now
appearing~\cite{msnfm2024,mgpc2026}. A skull with a defect is a partial shape and the implant
completes it, so implant design is close to what these models advertise~\cite{medshapenet2025}, and
a checkpoint used as it comes would remove both the training data and the manual step.

Completion models are judged by how closely the completed shape resembles a reference, and that
resemblance is not what a surgeon needs to know: it cannot say whether the model added anything the
patient's own bone did not already supply, nor whether what it predicted could physically be placed
in that patient~\cite{metricsreloaded2024,rankings2018,dcd2021}.

\textbf{Contribution:} We audit three publicly released pretrained shape completion models, applied
zero-shot to cranial defects, and judge them by what the clinic needs rather than by what a
benchmark reports. First, we score every output against an explicit do-nothing baseline,
the patient's own retained bone. Second, we introduce a physical acceptance criterion, under which
the predicted material must sit in the defect without intersecting the bone that remains, and we
report how many patients could receive each output. Third, we test whether the language
conditioning these models advertise carries anatomical information, reading each against its own
noise floor. Together they let a completion model be judged on the thing that decides its
use: whether what it produced could be put in a patient.

\section{Related Work}
\label{sec:related}

Automatic cranial implant design has been posed mainly as volumetric completion of the missing
bone. The AutoImplant challenges established the task over SkullBreak and SkullFix, whose defects
are carved synthetically from complete CT
skulls~\cite{autoimplant2020,autoimplant2021,skullbreak2021}, and networks trained for it handle
those defects well~\cite{li2021cranioplasty}; MUG500+ later added real craniectomy cases with their
manufactured implants~\cite{mug500}. These set the standard a zero-shot checkpoint has to meet.

Shape completion infers a complete shape from a partial observation~\cite{pcn2018}. We audit
every released model we could obtain that performs it without task training: the two offered at
or advertised at foundation scale, and one strong generalist as a non-medical control. AdaPoinTr is a strong pretrained generalist trained
on ShapeNet-55, whose 55 everyday object categories include no
skull~\cite{adapointr2023,shapenet2015}. MSN-FM describes itself as the first multi-modal
foundation model for medical point cloud completion and takes its text through a BERT
encoder~\cite{msnfm2024,medshapenet2025,bert2019}. MGPC, a more recent preprint, keeps a text branch
and adds an image branch, both read from frozen encoders~\cite{mgpc2026,clip2021,dinov2}; it
makes both modalities optional at inference, though its released code provides no such path, so an
image must always be supplied. MSN-FM is advertised as a foundation model and MGPC is trained at
that scale. Surveys of 3D point cloud foundation models reach neither of
them~\cite{pcfmsurvey2025}.

The benchmarks they were built on report similarity to a reference, which is a poor match for a
clinical decision. Chamfer distance, the usual choice, is insensitive to mismatched point density
and heavily influenced by outliers~\cite{dcd2021}, and without a do-nothing baseline no benchmark
separates a model from the trivial solution~\cite{metricsreloaded2024,reinke2024pitfalls}.
Completion trained on synthetic partials also degrades on real scans~\cite{pathak2025}. We
therefore fixed the baseline and the acceptance criterion before running a single model.

\section{Materials and Methods}
\label{sec:protocol}

We work with 239 skulls: 210 held-out test cases of SkullBreak (100) and SkullFix (110), whose
defects are carved synthetically, and 29 real craniectomy defects from
MUG500+~\cite{skullbreak2021,mug500}. The first two give the \emph{missing bone}, the whole removed
volume including the diploe; MUG500+ gives the \emph{manufactured implant} an expert designed for
that patient, so its results measure whether material went where the implant goes rather than
reconstruction accuracy.

Every model receives the kept subset of one surface sample of the \emph{complete} skull, split by
the missing-bone mask, so that input and reference come from a single draw. The median kept
subset holds 5955 points, reduced by farthest-point sampling to each architecture's fixed input
size: 4096 for MSN-FM, 2048 for the other two. We measure recall of the removed region at 2\,mm and report it as an
\emph{advantage} over a no-op, the per-skull difference against simply handing back the patient's
retained bone. That baseline is not optional: the removed bone is a thin shell over the vault, so
proximity to retained bone supplies coverage for free~\cite{metricsreloaded2024,rankings2018}.
The tolerance is 2\,mm because the task-specific model described below saturates at 5\,mm and
still reaches 0.966 at 2\,mm, so sparsity cannot excuse a coverage failure there.

Not every row spends its point budget the same way, so the headline table is not a like-for-like
comparison. A completion model spreads its points over a whole skull;
the task-specific reference emits the defect region alone, putting 2733 of its 6144 points inside
the removed region against 193 for an ideal uniform sample of a whole skull. That uniform resample is the
allocation-neutral row, and its median per-skull advantage is $+0.169$.
Recall rises with point count, so every arm is scored at a common 6144-point budget and at its
native one; the two conventions agree on the ordering and flip AdaPoinTr's sign ($+0.013$ native,
$-0.036$ common). Implantability is read at the common budget. Extra points can only help recall,
and because penetration is a maximum over points they can only hurt it, so the two conventions
pull in opposite directions.

A prediction is \emph{implantable} if it is collision-free at a 1\,mm tolerance against the
patient's retained bone \emph{and} its advantage over the no-op at 2\,mm is non-negative. That is
a geometric screen and a necessary condition rather than a clinical approval: an output failing it
cannot be placed, while one passing it still has to satisfy everything a designer and a surgeon
check. The tolerance is 1\,mm
because that is what the instrument resolves: on the 478 rows whose true penetration is zero by
construction it reads 0.533\,mm at worst. That is a measurement floor, not a surgical requirement,
and nothing in the audited models' verdict turns on it, their penetrations being ten times larger;
the reference row, at 0.55\,mm, is the one row it does bind. The second condition earns its place because collision-freedom alone is
gameable: one swept configuration returned a skull $0.634\times$ the patient's size, a median
26.4\,mm from any bone. Headline numbers use the \emph{deployment} frame, which places a model
from the defective skull alone; the \emph{training-time} frame aligns to the complete skull,
unobtainable when that skull is the prediction.

A task-specific volumetric model gives the criterion a scale: a 3D convolutional encoder--decoder
segments the missing bone and emits an implant mask, sampled to the same 6144 points and scored by
the same instrument. It was trained on the SkullBreak and SkullFix training splits,
disjoint from the test skulls, then selected as the best of 138 arms by Dice (0.916) on the same
210 carved cases it is reported on, so its row is a \emph{test-selected reference ceiling} and not
an unbiased estimate, and the 29 real defects are out of domain for it. We built it ourselves, and it
is here for one reason rather than as a rival: without it a reader could not tell whether these
criteria can be met at all.

The easier mistake here is to condemn a model that does not deserve it, so we spent the second
half of the study trying to break the first: we swept six protocol dimensions and added a positive
control handing each model the complete skull. One asymmetry is worth stating. MSN-FM's and MGPC's
checkpoints were accepted on reproducing upstream behaviour, AdaPoinTr's on a weaker check that the
weights had loaded, because the ShapeNet-55 test split is not available. The source code is released
at~\cite{code2026}.

\section{Results}
\label{sec:results}
\label{sec:rq1}

\begin{table}[tbp]
\caption{Medians over 239 skulls at the common 6144-point budget, deployment frame. Advantage is
the per-skull gain in recall of the removed region at 2\,mm over a no-op; \emph{implantable} adds
collision-freedom at 1\,mm. The controls pass by construction, not by merit, and the task-specific
row is ours and test-selected (\cref{sec:protocol}). 29 of the 239 are real craniectomy defects,
separated out in \cref{tab:clinical}.}
\label{tab:headline}
\centering
\footnotesize
\setlength{\tabcolsep}{4pt}
\begin{tabular}{@{}lrlrrc@{}}
\toprule
 & & & Adv. & Pen. & \\
Method & Params & Pretraining corpus & @2\,mm & (mm) & Impl. \\
\midrule
Ground truth       & --     & control                 & $+0.802$ & 0.35 & 239/239 \\
No-op, unchanged   & --     & control                 & $+0.000$ & 0.35 & 239/239 \\
Uniform resample   & --     & allocation-neutral      & $+0.169$ & --   & --      \\
\midrule
Task-specific volumetric & -- & trained on this task & $+0.745$ & 0.55 & 238/239 \\
\midrule
MSN-FM~\cite{msnfm2024}    & 297\,M & MedShapeNet, 100k       & $-0.156$ & 6.60 & 0/239 \\
AdaPoinTr~\cite{adapointr2023} & 32\,M & ShapeNet-55, 55 classes & $-0.036$ & 5.42 & 0/239 \\
MGPC~\cite{mgpc2026}       & 559\,M & MGPC-1M, 1M pairs       & $-0.130$ & 6.57 & 0/239 \\
\bottomrule
\end{tabular}
\end{table}

By the usual coverage score the models do not look broken. With the normalisation held fixed so that
only the defect process varies, MSN-FM recalls 0.462 of the removed region at 5\,mm on the
carved defects, against 0.439 on the compact defect it was trained to repair. Coverage alone cannot say whether a model added anything at all.

\begin{figure}[tbp]
  \centering
  \includegraphics[width=\textwidth]{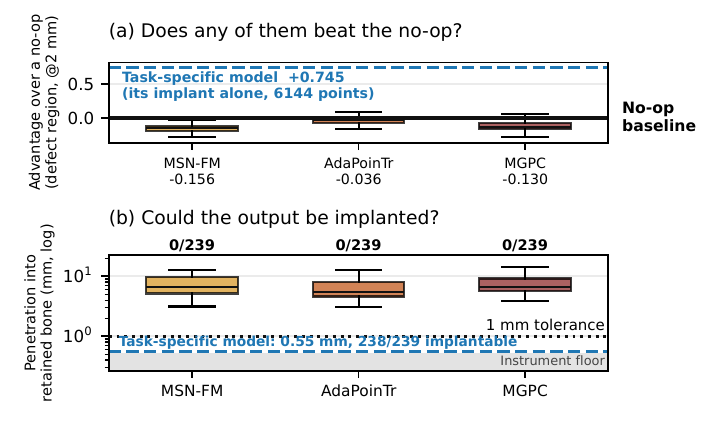}
  \caption{Three pretrained completion models, 239 skulls, equal 6144-point budget. (a) advantage
    over a no-op at 2\,mm; (b) deepest penetration into retained bone, log scale, implantable count
    above each box. In (b): dotted rule, the 1\,mm tolerance; dashed line, the task-specific
    reference, which is ours and test-selected (\cref{sec:protocol}); shaded band, the instrument's
    floor.}
  \label{fig:crossmodel}
\end{figure}

All three put less material within 2\,mm of the missing bone than the patient's own retained bone
already does (\cref{tab:headline}, \cref{fig:crossmodel}a): AdaPoinTr $-0.036$,
MGPC $-0.130$ and MSN-FM $-0.156$, where the task-specific reference reaches $+0.745$ on an
allocation that is not like-for-like with theirs (\cref{sec:protocol}). Medical pretraining did not
help: MSN-FM is the worst of the three, behind one trained on 55 everyday object categories.

The difference between a completion benchmark and a clinical defect lies in the baseline rather
than in the model. In that same comparison a model's coverage barely moves across defect types,
while a no-op recovers
0.049 of the compact ball-shaped defect used in completion training and 0.783 of the dispersed
defects the skull datasets ship, both at 5\,mm. What looks like competence on the released benchmark is mostly
the baseline.

\label{sec:reversal}
\begin{table}[tbp]
\caption{Advantage over a no-op at 2\,mm, median per skull, by defect provenance. The clinical
$p$-values are a two-sided sign test, counting AdaPoinTr's three tied skulls against it, and a
Wilcoxon signed-rank test, which drops them; they disagree for AdaPoinTr, whose clinical median is
therefore not established. Every synthetic result is significant at $p < 10^{-24}$ under both.
MUG500+ ships a manufactured implant, so its columns measure whether material went where the
device goes.}
\label{tab:clinical}
\centering
\footnotesize
\setlength{\tabcolsep}{3pt}
\begin{tabular}{@{}lrcrcrr@{}}
\toprule
 & \multicolumn{2}{c}{Synthetic ($n = 210$)} & \multicolumn{2}{c}{Clinical ($n = 29$)}
 & \multicolumn{2}{c@{}}{Clinical $p$} \\
\cmidrule(lr){2-3} \cmidrule(lr){4-5} \cmidrule(l){6-7}
Model & Adv. & Better & Adv. & Better & Sign & Signed-rank \\
\midrule
MSN-FM~\cite{msnfm2024}    & $-0.162$ & 0/210  & $-0.102$ & 1/29  & $<0.0001$ & $<0.0001$ \\
AdaPoinTr~\cite{adapointr2023}    & $-0.042$ & 29/210 & $+0.021$ & 16/29 & $0.7111$  & $0.0409$ \\
MGPC~\cite{mgpc2026} & $-0.139$ & 4/210  & $+0.118$ & 24/29 & $0.0005$  & $<0.0001$ \\
\bottomrule
\end{tabular}
\end{table}

Two things move the sign of the comparison with a no-op. At its native 8192-point budget AdaPoinTr
reaches $+0.013$, which the budget sweep attributes to the extra points rather than better
content. So does the cohort (\cref{tab:clinical}): on the 29 real craniectomy defects MGPC
beats a no-op at $+0.118$, on 24 of 29, while losing on the carved ones, and AdaPoinTr's $+0.021$
there is not established. One model reversing on 29 skulls, against a manufactured implant, does
not settle the clinical question.

\begin{figure}[tbp]
  \centering
  \includegraphics[page=1,width=\textwidth]{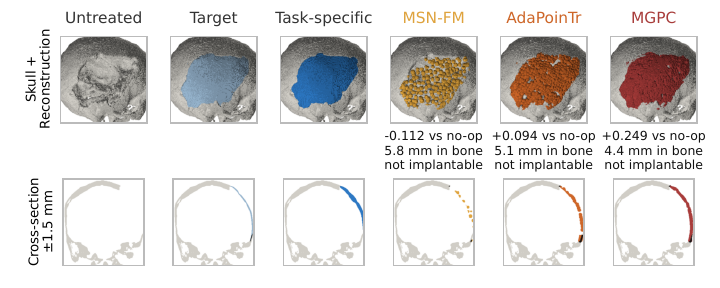}
  \caption{Case B0001 of MUG500+, from outside (top) and in section (bottom). The patient's own
    bone is grey, what each method adds is in colour, black marks material inside bone that is
    there. MSN-FM's addition is granular; AdaPoinTr's and MGPC's read as solid caps, though only
    AdaPoinTr's points lie on a surface. All three run into bone the patient still has. On this
    case two of them exceed the no-op; the medians in \cref{tab:headline} do not.}
  \label{fig:gallery}
\end{figure}

Seen on the patient all three cover the defect with material that reads as a completed skull
(\cref{fig:gallery}), but only one returns a surface: 0.564 of AdaPoinTr's points lie within 2\,mm
of the true skull, against 0.108 and 0.043 for the others.

\label{sec:rq2}
Neither kind can be placed in the patient. Every model is implantable on 0 of 239 skulls, against 238 of 239
for the task-specific reference, which is collision-free on all of them; across the three, not one of 2868 model-arm-skull combinations is
collision-free at 1\,mm (\cref{fig:crossmodel}b). Misregistration does not account for that: the unobtainable
training-time frame moves penetration by Cliff's $\delta = -0.025$ ($p = 0.109$) against a total
separation from the no-op ($\delta = 1.000$), so the collisions are in the shape of the output, not
in where it was put. Nor does the allocation: the ground-truth control spreads the same
6144 points over the same whole skull and collides on none of the 239 (\cref{tab:headline}).

\begin{figure}[tbp]
  \centering
  \includegraphics[width=\textwidth]{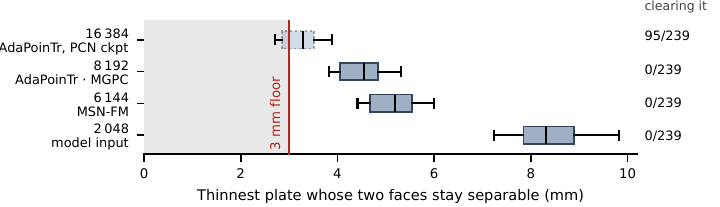}
  \caption{How thin a plate can get before its two surfaces stop being distinguishable, at each
    point budget in play (ideal even resample, no model, 239 skulls per box).
    No budget the audited models emit or receive reaches 3\,mm, so thickness is lost before a
    model is asked to recover it. That
    3\,mm is a design target rather than a law, and the clinical data break it: 11 of the 29 real
    MUG500+ implants (38\,\%) are thinner than 3\,mm, the 29 having median 3.05\,mm and mean
    $3.30\pm1.41$\,mm. The dashed box is AdaPoinTr's other checkpoint (\cref{sec:robustness}).}
  \label{fig:identifiability}
\end{figure}

At the point densities these models emit, implant thickness is not carried by the output at all,
whatever the model (\cref{fig:identifiability}). A 6144-point sample of a skull has a mean nearest-neighbour spacing
of 2.56\,mm, and at a comparable spacing of 2.4\,mm, 0.447 of every neighbourhood on a 3\,mm plate is drawn
from the opposite face, so the two faces are not separable and thickness cannot be read off the emitted
points~\cite{amenta1999}. Even a 5\,mm plate is contaminated at 0.351, whereas at 0.4\,mm
spacing it separates cleanly, which places the failure in the representation and not in the
measurement. The 3\,mm floor is itself a design
target~\cite{fishman2024thickness} that the clinical data violate, 11 of the 29 real MUG500+
implants being thinner than it (\cref{fig:identifiability}).

\label{sec:rq3}
The other advertised capability is language, which only the two foundation models accept. Against
each model's own no-information floor, every MGPC prompt moves the geometry 0.05 to 0.15$\times$ as
much as a second blank image, and MSN-FM's 0.49 to 1.08$\times$ as much as a change of seed, so
most of MSN-FM's prompt swaps do less than asking it twice. MSN-FM's weights say why: its text branch ends in a ReLU, 97.5\,\% of the text code is exactly
zero, and a prompt moves the fused code by 8.8\,\% of the shape code's norm. Across two text
encoders and two fusion mechanisms the channel is decorative in MGPC and, on all but one prompt,
in MSN-FM.

\label{sec:robustness}
None of this rests on protocol choices of ours. Presentation pose and input scale showed no effect on a
14-skull pilot, which at that size is weak evidence, and we do not read it as evidence of no
effect. Across all
six swept dimensions, no audited row is implantable. The one arm that
comes closest is instructive: AdaPoinTr's other released checkpoint emits 16\,384 points, twice the
largest audited budget and the only one that resolves a 3\,mm plate
(\cref{fig:identifiability}), and it is collision-free on 1 of 239 skulls. It is also the worst arm
we ran, at $-0.198$. Having enough points to represent an implant is not the same as producing one. Handed the
complete skull, AdaPoinTr clears a no-op at $+0.042$ and improves on its own default on 228 of 239
skulls, so a negative number is no artefact of our frame or input. MSN-FM ($-0.151$) and MGPC
($-0.124$) cannot return even a shape they were shown.

\section{Discussion}
\label{sec:discussion}
\label{sec:conclusion}

The models answer a different question from the one implant design asks, and thickness shows it.
Measuring it at all means turning an output into a solid, which we do with a 3\,mm reconstruction
radius, identically for every method. What comes back barely moves: 5.40 to 5.93\,mm for every
model on every cohort, against references spanning 3.05 to 7.25\,mm. The number is our reconstruction
radius rather than their output, and that is the point: a surface has no thickness, so there is
nothing for the measurement to find, and no overlap metric here registers that.

Two barriers stand behind that, and neither would move for a loss weight or more data. The first
is that the corruption these models were trained to repair is a different object from a clinical
defect: a compact cluster removing 10.8 times as many surface points and differing in dispersion at
Cliff's $\delta = 0.83$. The second is the output type: a surface has no thickness to recover, at a
density that could not resolve one (\cref{fig:identifiability}).

How the failure presents is what makes it a trust problem. Look at the pictures and the models have
done their job: the defect is covered, the head has its shape back, and AdaPoinTr returns something
a reader would accept as a skull. Measure the same outputs and none can be put in the patient. A
diffuse cloud is its own warning; a clean cap that scores unremarkably and still cuts into bone
carries none, and that is the harder case to catch.

The outputs are not beyond repair. A cloud sitting roughly in the right place could be surfaced,
trimmed against the retained bone and thickened into a plate. But each step carries a threshold
whose right value differs from skull to skull, so they would be set by hand, per patient. That is
the manual work a pretrained model was meant to remove, and a pipeline needing it is not automatic. What it has to
beat is the time a specialist spends in CAD, and repairing these completions is unlikely to be
quicker.

Two changes follow, one to the models and one to how they are judged. Recovering thickness needs a
volumetric or implicit output, or point spacing below half the plate thickness, and a training
defect that is the clinical one. Physical acceptance has to enter the objective, and any overlap
score needs a do-nothing baseline~\cite{metricsreloaded2024,reinke2024pitfalls}.

What this does not cover is worth saying plainly. We audited released checkpoints used as they
come, so what finetuning on cranial data would buy is untested here. The acceptance criterion is
geometric and no clinician read the outputs: failing it is disqualifying, passing it would not by
itself make an implant acceptable. We also did not run a task-specific point-cloud method, which
would separate the output representation from the
pretraining~\cite{friedrich2023,kwarciak2026pcfm}. And the real defects number 29, enough to show
that carved benchmarks flatter these models and not enough to settle a clinical question alone.

To conclude, cranial implant design looks, at first sight, like a shape completion problem. That
assumption should be treated with caution: the released shape completion foundation models we could
obtain are not trustworthy here, and an output can look like a correct reconstruction and still be
of no use in the clinic. Research towards trustworthy foundation models for medical shape
completion remains at an early stage.

\subsubsection*{\ackname}
The project was funded by The National Centre for Research and Development, Poland under Lider
Grant no: LIDER13/0038/2022 (DeepImplant). We gratefully acknowledge Polish high-performance
computing infrastructure PLGrid (HPC Center: ACK Cyfronet AGH) for providing computer facilities
and support within computational grant no. PLG/2026/019392.

\FloatBarrier
\label{paper:lastbodypage}

\bibliographystyle{splncs04}
\bibliography{main}

\end{document}